# Dynamics of core of language vocabulary

Valery D. Solovyev, Vladimir V. Bochkarev, Anna V. Shevlyakova

Kazan Federal University
maki.solovyev@mail.ru

**Abstract.** Studies of the overall structure of vocabulary and its dynamics became possible due to creation of diachronic text corpora, especially Google Books Ngram. This article discusses the question of core change rate and the degree to which the core words cover the texts. Different periods of the last three centuries and six main European languages presented in Google Books Ngram are compared. The main result is high stability of core change rate, which is analogous to stability of the Swadesh list.

## 1 Introduction

In this paper, we investigate the dynamics of the overall structure of the language vocabulary from a cognitive point of view. Traditionally, two components of the language vocabulary are distinguished: the center and periphery. The former contains highly stable words of maximum frequency (go, read, etc.) and provides stability to the language; the periphery contains the words that have become outdated or, on the contrary, have just appeared in the language, and thus, guarantees greater flexibility to it. We will present some quantitative characteristics of the dynamics of the center.

To do it, we should answer the following questions. How to determine the core? What is the size of the core? What is the rate of change of the core? What is the overall frequency of the core words? We will refer to Google Books Ngram corpus to answer these questions (https://books.google.com/ngrams). Similar problems were considered in [1, 2]. The frequency approach is a standard approach used to study core formation. In this paper, we consider two kinds of frequency: the word occurrence frequency in the corpus and the share of books in which the word occurs. Though these approaches are rather close, yet there are some differences.

The first question to answer is how to determine the core. It's impossible to define a clear boundary of the core. For example, the known Swadesh wordlists contain 40, 100 or 200 items. In [1] the core contains 100 words. It appears to be too limited. Let us note that Basic English contains 850 words, and the basic set of root words of Esperanto contains 900 items. The Voice of America's Special English [3] and Wikipedia in Simple English use, correspondingly, about 1500 and 2000 words. The basic vocabularies for foreigners [4], creole [5] and pidgin languages [6] contain 1.5 to 3 thousand words.

In [2] the core is composed of 1000 most frequent words (the first 100 words constitute what is called the head, and words 101 to 1000 form the body), and the periphery consists of the following in frequency 6000 words. In [9] the size of core vocabulary that provides a specified percentage of word usage based on the Google Books Ngram data is calculated. Thus, 2300 most frequently used English words have the total relative frequency of 75 %.

We carry out calculations not only for one fixed core, but for consecutive variants: for 1000, 2000, …, 8000 most frequent words, covering the whole range described above.

The following data preprocessing which allowed reducing the number of mistakes in the used data base was performed in this work. Only lexical 1-grams were selected which consisted only of the corresponding alphabet letters and one apostrophe in some cases. To normalize and calculate the relative frequencies, the number of lexical 1-grams was calculated for each year (as distinct from the Google Books

Ngram Viewer where the normalization is made for the total number of all 1-grams). Parts of speech are marked in the 2012 version of the corpus. But parts of speech are marked wrongly in many cases which can result in incorrect conclusions based on these data. We used the method explained in [9], i.e. if the number of word forms corresponding to some part of speech doesn`t exceed 1 % of total frequency of the given word form, such word forms were marked and not used during further analysis.

## 2    Rate of change of the core

When considering the rate of change of the core, we calculate the share of words of the core excluded from it during a given period. Figure 1 shows the relevant data for an interval of 50 years in English language. Changes of word frequencies can be due to both language evolution and random factors. To eliminate these factors, frequencies of word usage were studied throughout rather long 50-year intervals: 1676-1725, 1726-1775, … 1976-2008. Then, the words were ranked in decreasing frequency order and the percentage of words, which dropped out from the core of the successive 50-year interval were calculated. For example, the columns of the diagram marked "1825" show the percentage of core words for the period 1776-1825 which dropped out from the core in 1826-1875.

We observe a rather steady rate of updating of the core in the last 300 years: an average of 13-15% of the words drop out of the core in 50 years. Of course, it does not mean that these words disappear from language, only their frequency decreases, and they are forced out from the core by other words. There is not enough data in Google Books Ngram for the previous period (1500–1700), and therefore they are not provided here. Curiously, the updating rates of the core decrease during the Victorian era and increase in the first half of the 20th century. Also, it should be noted that the found mean value of 13-15% almost does not depend on the core size in the range from 1 to 8 thousand.

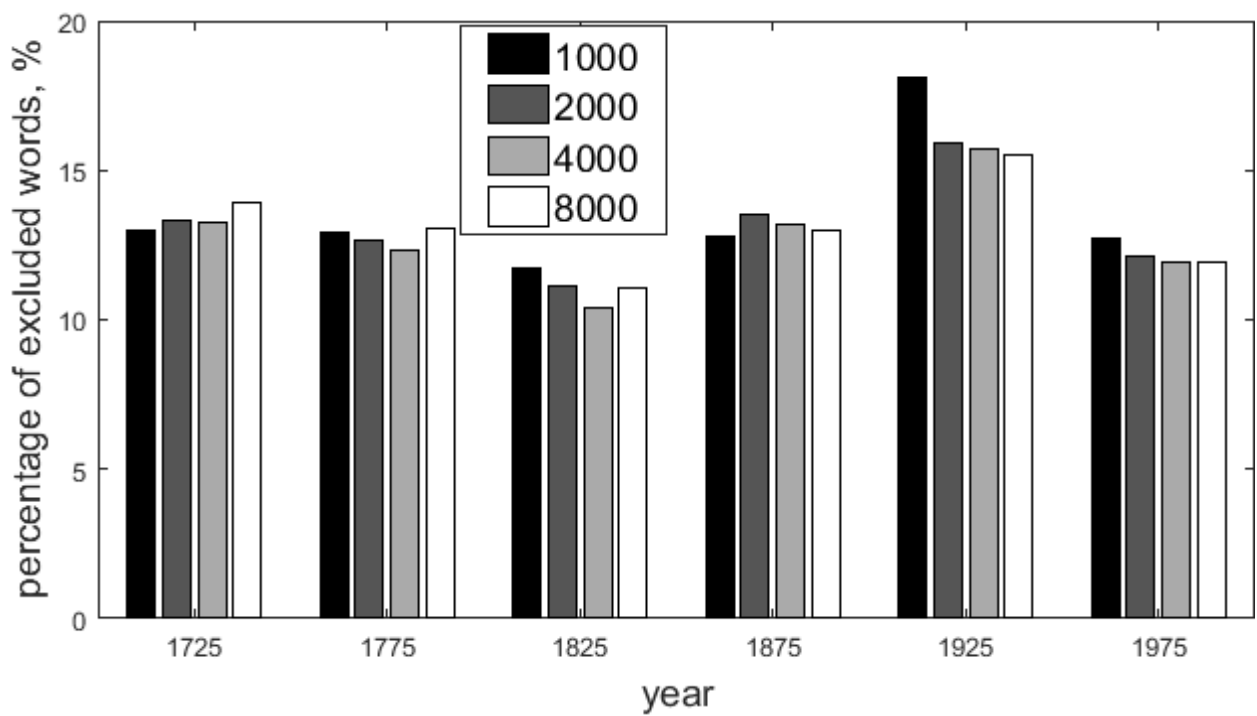

**Fig. 1.** Share of English words dropped out of the core in a 50-year period

When the core is defined through the share of books, the following changes occur in its content. If we select all English words that are found at least in one out of two books, we obtain a wordlist of 2302 items. We can construct, for comparison, a list with the same quantity of most frequent words. In spite of the fact that the share of books in which a word is used correlates poorly with its frequency (the correlation coefficient for all words of English language is just 0.15, for one thousand of the most frequent words it is 0.25), both lists overlap by 79%. At the same time, the differences between the lists are quite essential – there are 482 words that appear just in one list.

Words included in list 1 seem to be, according to the intuitive perception of the language, the most suitable for the core group of words. List 2 contains words that can hardly be attributed with certainty to the core vocabulary. These words correspond, first of all, to geo-graphical names and vocabulary with related meaning (for example, Africa, African, Rome, Berlin, Japan, Japanese, Spain, Spanish, India, Indians, Canada, California, Virginia, Asia), proper names/appellations (Wilson, Richard, Louis, Oxford), parts of words/letters that entered the list accidentally (ff), abbreviations (cf, vol., al, ibid.), articles and prefixes in loanwords and found foreign vocabulary (der, des, du, le, les, un, el), words belonging more to professional vocabulary than to common (carbon, oxygen, copper, equation, electron, protein), loanwords (bureau), words connected chiefly with political actions (socialist, colonial, empire, queen). However, according to the intuitive notion of language core, it is difficult to ascribe the words from the specified groups to the core, but we should not deny their importance for English-speaking society. In the culturological context, the words Oxford and queen for British people are undoubtedly important, as well as the words California, Africa and Virginia for Americans; additionally, professional words come into broad use together with the growth of public aware-ness.

As for the dynamics of the core (updating by 13-15% in 50 years), it practically does not change, regardless of these two ways of determination.

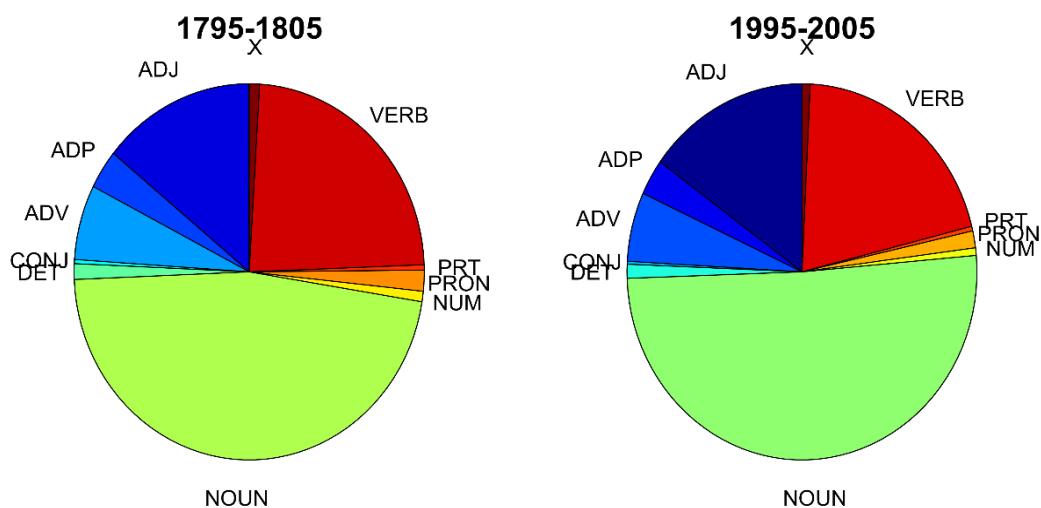

**Fig. 2.** Shares of various parts of speech in the 2000-core 200 years ago and today

Let us consider the structure of the core from the perspective of the parts of speech. In the latest version of Google Books Ngram, English words have been marked as parts of speech with 95% accuracy [10]. In figure 2 we can see the share of each part of speech around the year 1800 and today.

X stands for abbreviations, foreign words or words whose membership to a part of speech has not been determined. In 200 years the share of nouns and verbs has diminished. Figure 3 shows the dynamics of the parts of speech. The algorithm for marking the parts of speech works with higher accuracy in the case of modern words; this is why the share of X is the one declining most rapidly.

As one would expect, the parts of speech with the highest content, i.e. nouns and verbs (about 45%), drop out at the highest rate, while auxiliary parts of speech, articles, conjunctions, etc. (about 15 to 20%), do it at the lowest rate.

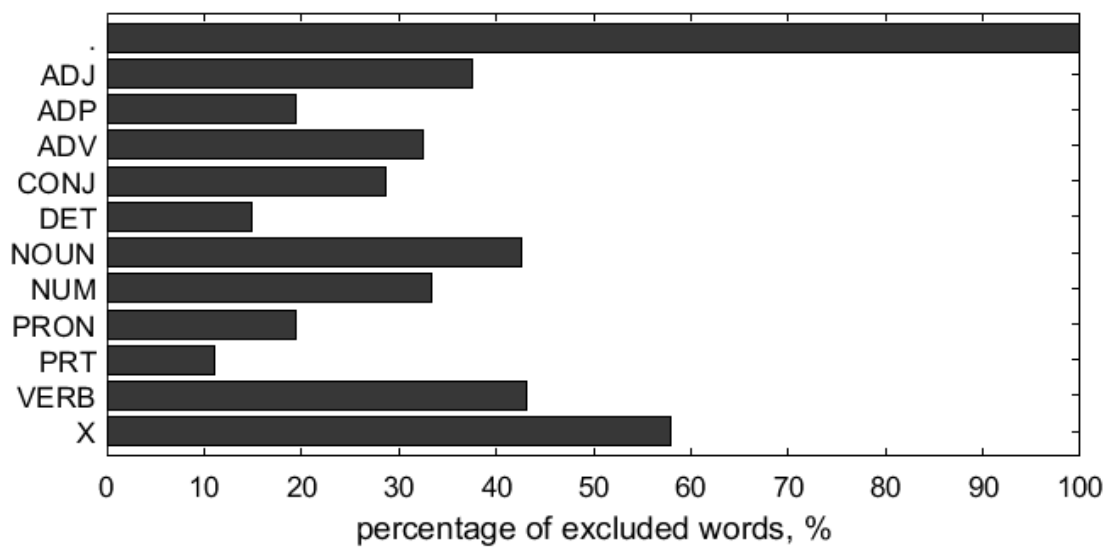

**Fig. 3.** Shares of various parts of speech dropped out of the 2000-core in 200 years

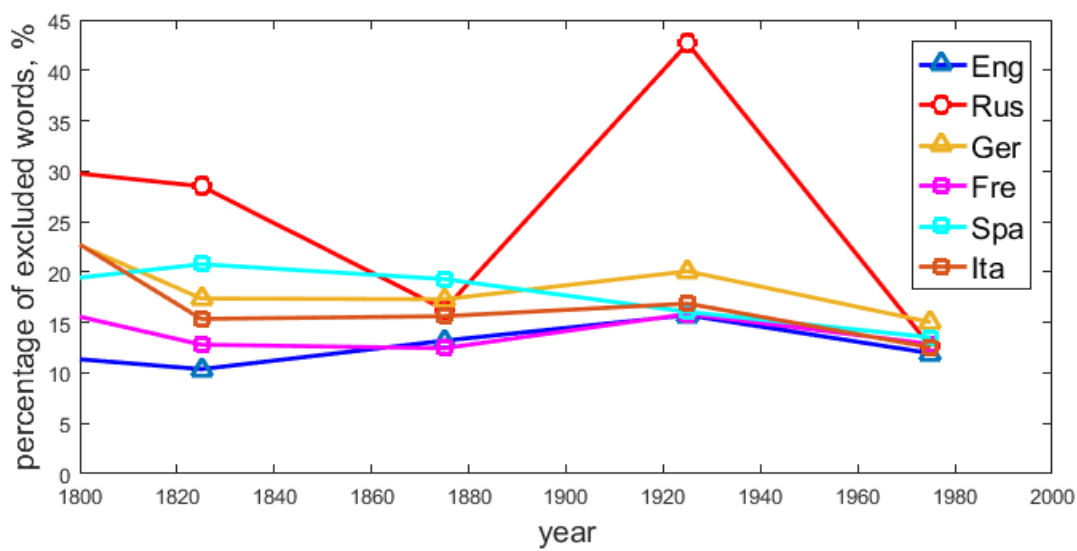

**Fig. 4.** Dynamics of the core (4000 words) for major European languages

Similar data are obtained for the main European languages (fig. 4) representing three different branches of Indo-European languages: Slavic, Romance and German, which separated just a few thousand years ago. This is somewhat similar to Swadesh results. Russian rather stands out from the general picture. The social upheavals in the beginning of the 20th century (the socialist revolution, which led to radical economic, political, cultural changes) were reflected in the vocabulary core.

## 3   Degree of covering of texts by the core

The important characteristic of core words is to what extent they are efficient for communication. Formally, this can be presented by percentage of core words in the texts, in other words by the degree to which the core words cover these texts. Let us analyze now the change of the total frequency of words of the core, that is the degree of covering of texts by these words. If one considers the core for the language state in 1800 (for a higher stability in calculations one takes the interval 1795–1805 and defines the core in the whole interval), it is evident that some words from the core will become outdated, and the overall frequency will fall over time. The exact quantitative characteristics of this process are given in figure 5 (the left window).

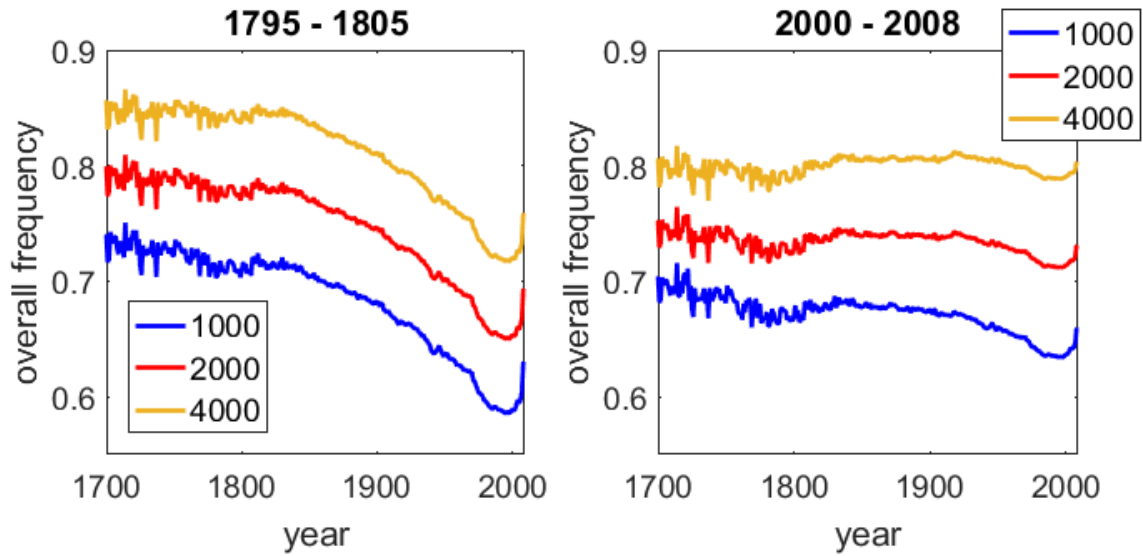

**Fig. 5.** Dynamics of the overall frequency of core words for the year 1800 (on the left) and 2000 (on the right) if the core size is different

For a 1000-word core the overall frequency falls in 200 years approximately from 0.7 to 0.6. Frequency curves for cores of bigger sizes look similarly. This effect may be explained not only by the obsolescence of the words of the core (their removal from the core), i.e. by the up-dating of the language, but also by the extension of the vocabulary, which in general grants greater expressive opportunities to the language and, naturally, leads to the reduction of the share of old words. According to data provided in [7], the number of words in English language grew from 544,000 in 1900 up to 1,022,000 in 2000, i.e. almost twice.

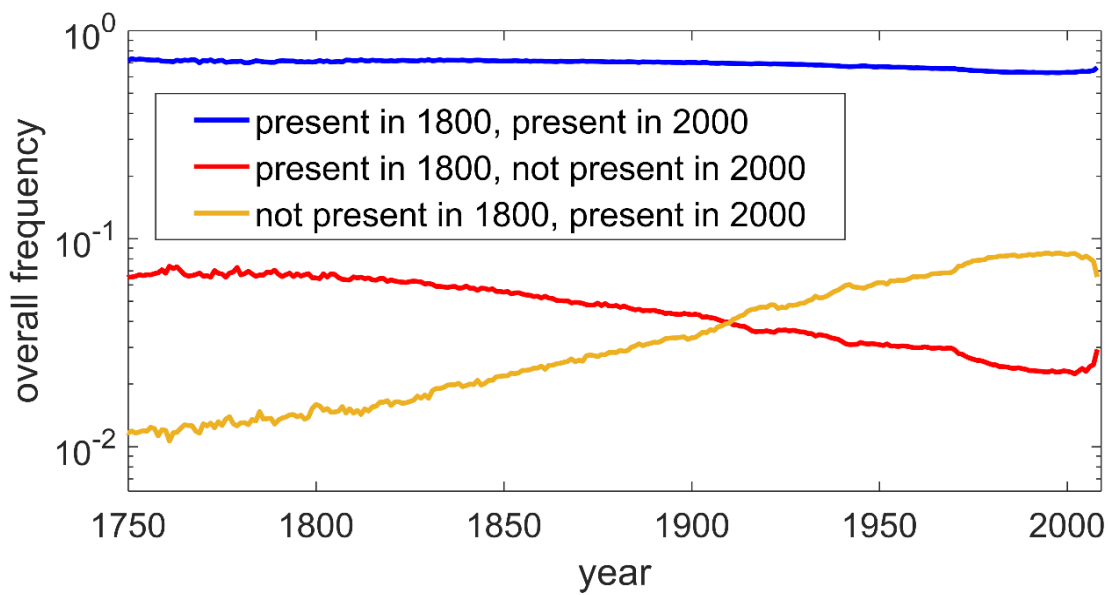

**Fig. 6.** Dynamics of total frequencies of various groups of words in the 1800 and 2000 cores

If one considers the modern core (years 2000–2008), the dynamics of its frequency looks as follows (fig. 5, the right window). Here two tendencies confront. On the one hand, it is evident that two hundred years ago the frequency of modern words was lower (up to 0), and it seems that one should expect a growth in the frequency of these words. But, on the other hand, as we see in the previous diagram, the frequency of words of the core in general falls. And these two tendencies approximately counterbalance each other. The overall frequency of words for a core with 4 thousand words remains at the level of approximately 0.8, for a core of 1000 words it slightly falls from 0.67 to 0.65. The next graph (fig. 6) explains the essence of the processes taking place. Here we can see separately the words that are present in the core both in 1800 and in 2000, and also the words present in one of them but not in the other.

The overall frequency for the words remaining in the core during these two centuries decreases from 0.7 to 0.6. The frequency of the words that drop out of the core decreases, and that of the words entering the core increases, and this augmentation is more intensive than the loss of frequency of the previous group.

These data must be taken into account when analyzing the frequency dynamics of different groups of rather-high-frequency vocabulary. The frequency dynamics of basic emotions are studied in [8]. Data for English are presented in figure 7 (taken from [8]). One can see that the overall frequency of emotive vocabulary considerably decreases from 1800 to 2000. A priori this can be explained either by a reduction of emotionality of people (or at least that of texts) during this period, or by a general reduction of the frequency of all the words of the core, which includes also the considered emotive words. Comparison of the frequencies shows that the main acting factor is the first one. The frequency of emotive vocabulary decreased approximately by 50%, while the overall frequency of the words of the whole core decreased just by 15%. Thus, the reduction of the frequency of emotive vocabulary cannot be explained only by the reduction of the frequency of the whole core.

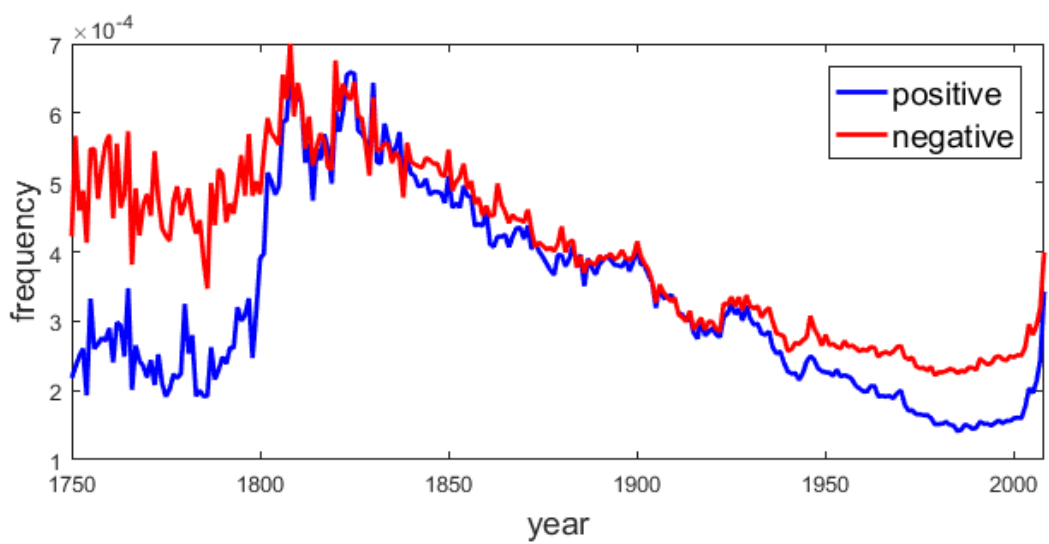

**Fig. 7.** Dynamics of total frequency of English emotive vocabulary [8]

## 4    Conclusion

In the article, the lexicon structure is considered from cognitive point of view distinguishing the center (core – the most frequently used lexis) and periphery. The core size is evaluated differently in different papers – from 1 to 8 thousand words. In our paper, the calculations are performed for all core sizes in this range. The core change data are presented for the first time. It turned out that the core has steadily changed during the last 300 years – approximately 15% of words is substituted every 50 years. The result is obtained for different languages (which are presented in Google Books Ngram) and is, to some extent, analogous to the results obtained by Swodesh concerning the stability of words from his list. The size of texts covered by the core words is counted (or the total frequency of core words). It was found that the core

(for the contemporary language) consisting of 1 thousand words covers two thirds of texts. If we regard the core words in 1800, the share of texts covered by them decreases from 0.7 to 0.6 for the last 200 years. This effect can be explained not only by core words obsolescence (removing from the core), i.e. by language updating but also by lexicon expansion which offers significant expression opportunities to a language and results in decreasing of old words percentage.

**Acknowledgements.** This research was supported by the Russian Foundation for Basic Research (grant № 15-06-07402).